\documentclass{ecai} 

\usepackage{times}
\usepackage{soul}
\usepackage{url}
\usepackage[hidelinks]{hyperref}
\usepackage[utf8]{inputenc}
\usepackage{subcaption}
\usepackage{graphicx}
\usepackage{amsthm}
\usepackage{algorithm}
\usepackage{algorithmic}
\usepackage[switch]{lineno}

\usepackage{newfloat}
\usepackage{listings}

\usepackage{booktabs}       
\usepackage{makecell}
\usepackage{amsmath,amssymb}
\usepackage{bm}
\usepackage{placeins}
\usepackage{multirow}
\usepackage{adjustbox}
\usepackage{amsfonts,dcolumn}
\usepackage[capitalize]{cleveref}

\newcommand{\BibTeX}{B\kern-.05em{\sc i\kern-.025em b}\kern-.08em\TeX}

\begin{document}


\begin{frontmatter}


\paperid{314} 


\title{TransFeat-TPP: An Interpretable Deep Covariate Temporal Point Processes}


\author[A]{\fnms{Zizhuo}~\snm{Meng}}
\author[A]{\fnms{Boyu}~\snm{Li}}
\author[B]{\fnms{Xuhui}~\snm{Fan}}
\author[A]{\fnms{Zhidong}~\snm{Li}}
\author[A]{\fnms{Yang}~\snm{Wang}}
\author[A]{\fnms{Fang}~\snm{Chen}}
\author[C,D]{\fnms{Feng}~\snm{Zhou}\thanks{Corresponding Author. Email: feng.zhou@ruc.edu.cn}}

\address[A]{Data Science Institute, University of Technology Sydney, Australia}
\address[B]{School of Computing, Macquarie University, Australia}
\address[C]{Center for Applied Statistics and School of Statistics, Renmin University of China, China}
\address[D]{Beijing Advanced Innovation Center for Future Blockchain and Privacy Computing, China}


\begin{abstract}
The classical temporal point process (TPP) constructs an intensity function by taking the occurrence times into account. 
Nevertheless, occurrence time may not be the only relevant factor, other contextual data, termed covariates, may also impact the event evolution. Incorporating such covariates into the model is beneficial, while distinguishing their relevance to the event dynamics is of great practical significance. 
In this work, we propose a Transformer-based covariate temporal point process (TransFeat-TPP) model to improve the interpretability of deep covariate-TPPs while maintaining powerful expressiveness. TransFeat-TPP can effectively model complex relationships between events and covariates, and provide enhanced interpretability by discerning the importance of various covariates. Experimental results on synthetic and real datasets demonstrate improved prediction accuracy and consistently interpretable feature importance when compared to existing deep covariate-TPPs. 
\end{abstract}

\end{frontmatter}


\section{Introduction}
\label{intro} 

A temporal point process (TPP) is a mathematical framework used to describe the occurrence of events in time. It is a statistical model that is widely used in fields such as ecology~\cite{law2009ecological,warton2010poisson}, epidemiology~\cite{gatrell1996spatial,meyer2010spatio}, neuroscience~\cite{paninski2004maximum,zhou2022efficient} and telecommunication~\cite{daley2003introduction,jahnel2020probabilistic}. In a TPP model, events are represented as points in time, and the goal is to understand the underlying process that generates these events. The model can be used to predict the occurrence of future events, identify patterns or trends in the data, and estimate the probability of different outcomes. Classical TPPs include the Poisson process~\cite{kingman1992poisson}, Cox process~\cite{moller1998log} and Hawkes process~\cite{hawkes1971spectra}, etc. 
These models are typically formulated in a parametric manner and often suffer from limited expressiveness. To overcome these limitations, deep TPPs leverage deep learning architectures and tools such as RNN~\cite{du2016recurrent,mei2017neural,shchur2019intensity}, auto-grad in DNN~\cite{omi2019fully}, self-attention mechanism~\cite{zhang2020self, zuo2020transformer} to more effectively model event generation processes.


However, an inherent flaw of both parametric and deep TPPs is that they cannot include additional information about relevant variables that may influence the occurrence of the events. On the contrary, a \emph{covariate}-TPP~\cite{reinhart2018self,truccolo2005point} is a type of TPP that incorporates additional information, called covariates, that may be related to the event occurrence. Covariates can be any relevant information, such as weather conditions, demographic characteristics or physical attributes of the location. It is expected that incorporating such covariates into events modelling can be more beneficial than considering events only, because by including covariates the model can account for the relationship between the events and other factors that may affect them. As far as we know, the research on covariate-TPPs is currently limited compared to the extensive results on TPPs. 

Due to the rapid development of deep learning, the \emph{deep} covariate-TPPs has emerged as a burgeoning topic, which refers to a type of covariate-TPPs that incorporates deep learning techniques to model complex relationships between events and covariates. For example, \cite{xiao2017intensity} integrates diverse time-series signals and events using two parallel RNNs, allowing for the joint modelling of event dynamics; \cite{okawa2019deep} presents a deep mixture point process model that leverages rich, multi-source contextual data to define the intensity function using a combination of kernels. 
While aforementioned works can be flexible and powerful in modelling such relationships, the resulting model may be difficult to interpret and explain because neural networks are inherently black box in nature. This lack of interpretability can make it challenging for researchers to gain insights from the model and draw meaningful conclusions from the results. Moreover, existing covariate-TPPs works potentially make a strong assumption that the covariates to be incorporated always hold a strong relationship with the events' evolution. For instance, \cite{xiao2017intensity} incorporates the ATM's models, age, and location (covariates) to assist in the modelling of ATM's errors or failures (events). 
However, the selection of covariates always requires extra expertise to determine whether the covariates hold a strong relationship with the events of interest. 
Offering further insights about the covariates' effect on events can significantly improve the interpretability of covariate-TPPs but also remains an open challenge. In summary, while deep covariate-TPPs offer better expressiveness, their interpretability remains limited and researchers must balance these tradeoffs when using this approach.

To enhance the interpretability of deep covariate-TPPs while preserving their expressive power, 
we propose a \textbf{Trans}former-based covariate-\textbf{TPP} model equipped with \textbf{feat}ure importance (TransFeat-TPP). The Transformer is employed to encode sequences and model the relationships among events. 
Owing to its highly parallelizable architecture, the Transformer enables simultaneous computations, resulting in substantial improvements in training and inference speed compared to conventional RNN models \cite{du2016recurrent, mei2017neural, omi2019fully, xiao2017intensity}. 
Furthermore, the Transformer's self-attention enables expressive input representations, addressing vanishing gradient issues in sequential models and enhancing long sequence handling.
Most importantly, our model includes a module called the feature importance self-attention neural network (Fi-SAN). Fi-SAN, which is originally introduced in \cite{vskrlj2020feature}, acts as a classifier to predict the next event type while also unsupervisedly learning the importance of different features, allowing it to identify and prioritize the most relevant ones. 

Specifically, our contributions are as follows:
\textbf{(1)} We introduce TransFeat-TPP, a model designed to integrate supplementary covariates, which can provide valuable insights into the evolution of events. The Transformer architecture allows it to effectively model complex relationships between events and covariates. 
\textbf{(2)} In TransFeat-TPP, we incorporate the Fi-SAN module to discern the correlation between covariates and the occurrence of events. Unlike existing deep covariate-TPPs, our model can provide the importance of various covariates, thus offering enhanced interpretability. As far as we know, this work should be the first deep covariate-TPPs with covariate ranking. 
\textbf{(3)} We validate our model on both synthetic and real datasets. Experimental results demonstrate that our model achieves a higher prediction accuracy compared to traditional TPPs. More importantly, in contrast to existing deep covariate-TPPs, our model can consistently provide interpretable feature importance. 

\section{Related Work}
In this section, we primarily review recent developments in the areas of TPPs, covariate-TPPs and feature importance. 

\subsection{Temporal Point Processes}

Various models for TPPs have been proposed. Early statistical models include the Poisson process~\cite{kingman1992poisson} and the Hawkes process~\cite{hawkes1971spectra}, which laid the foundation for understanding event occurrences. The Poisson process assumes independent events, while the Hawkes process allows for event dependencies through self-excitation. In recent years, deep learning models have become popular for TPPs, offering improved prediction performance and modelling complex relationships. Examples include the recurrent marked temporal point process~\cite{du2016recurrent}, neural Hawkes process~\cite{mei2017neural}, fully neural network point process~\cite{omi2019fully}, and Transformer Hawkes process~\cite{zuo2020transformer}, among others. These models leverage deep learning's expressive capabilities to capture temporal dependencies and enhance prediction accuracy in event sequences. However, the introduction of deep neural network may lead to some interpretability challenges due to its black-box nature.

\subsection{Interpretability in Temporal Point Processes}
Interpretability in temporal point processes remains an open challenge and has consistently attracted research interest in recent years. Explaining why events happen and revealing the causalities behind them is one aspect of interpretability. For example, temporal logic point processes \cite{pmlr-v119-li20p, li2022explaining} utilize time logic rules to incorporate domain knowledge, aiming to elucidate the causality behind event dynamics. Additionally, some researchers focus on revealing the patterns of influence among event types, which are often explicitly learned through existing popular deep TPPs \cite{du2016recurrent, mei2017neural,zhang2021learning, zuo2020transformer}. Studies such as \cite{shou2023influence, zhang2021learning} use graphical neural networks or self-attention mechanisms to explore these influencing patterns among event types.
However, the aforementioned studies primarily focus on traditional TPPs, which consider only the event information. When it comes to covariate-TPPs, the scope of interpretability significantly broadens, presenting a rich area for further exploration and development. In this work, our focus on interpretability primarily revolves around revealing the relevance of covariates to the event dynamics. 

\subsection{Covariate Temporal Point Processes}

Covariate-TPPs have garnered significant attention due to their ability to model the impact of external factors on event sequences. 
To the best of our knowledge, current research either incorporates covariates as a factor in the intensity function~\cite{adelfio2021including, okawa2021dynamic} or combines them with event sequences through deep neural networks~\cite{solomon2022deep,xiao2017intensity}. These models enabled researchers to analyze and predict event sequences more effectively, taking into account the various factors that influence event occurrences. 
However, treating covariates as a fixed factor in the intensity function explicitly results in reduced flexibility and expressiveness. Moreover, existing deep covariate-TPPs lack interpretability concerning the involvement of covariates. All these methods are based on the assumption that all covariates have a significant relationship with the events. This assumption requires domain expertise and may not always be valid in real-world scenarios.

\subsection{Feature Importance}
Feature importance plays a crucial role in interpreting machine learning models, particularly in applications where identifying relevant features is vital for decision-making and gaining insights. Various techniques have been developed to measure feature importance, ranging from traditional statistical methods like correlation coefficients~\cite{weisstein2006correlation} and information gain~\cite{kent1983information} to more advanced methods like permutation importance~\cite{altmann2010permutation} and LASSO regularization~\cite{zou2006adaptive}. In recent years, attention mechanisms
have emerged as powerful tools for estimating feature importance. Fi-SAN~\cite{vskrlj2020feature} leverages self-attention to learn the significance of different features in an unsupervised manner. By assigning attention weights to each feature, the model can effectively identify the most relevant features and prioritize them during the learning process.

\section{TransFeat-TPP}

\begin{figure}
    \centering
    \includegraphics[width=\linewidth]{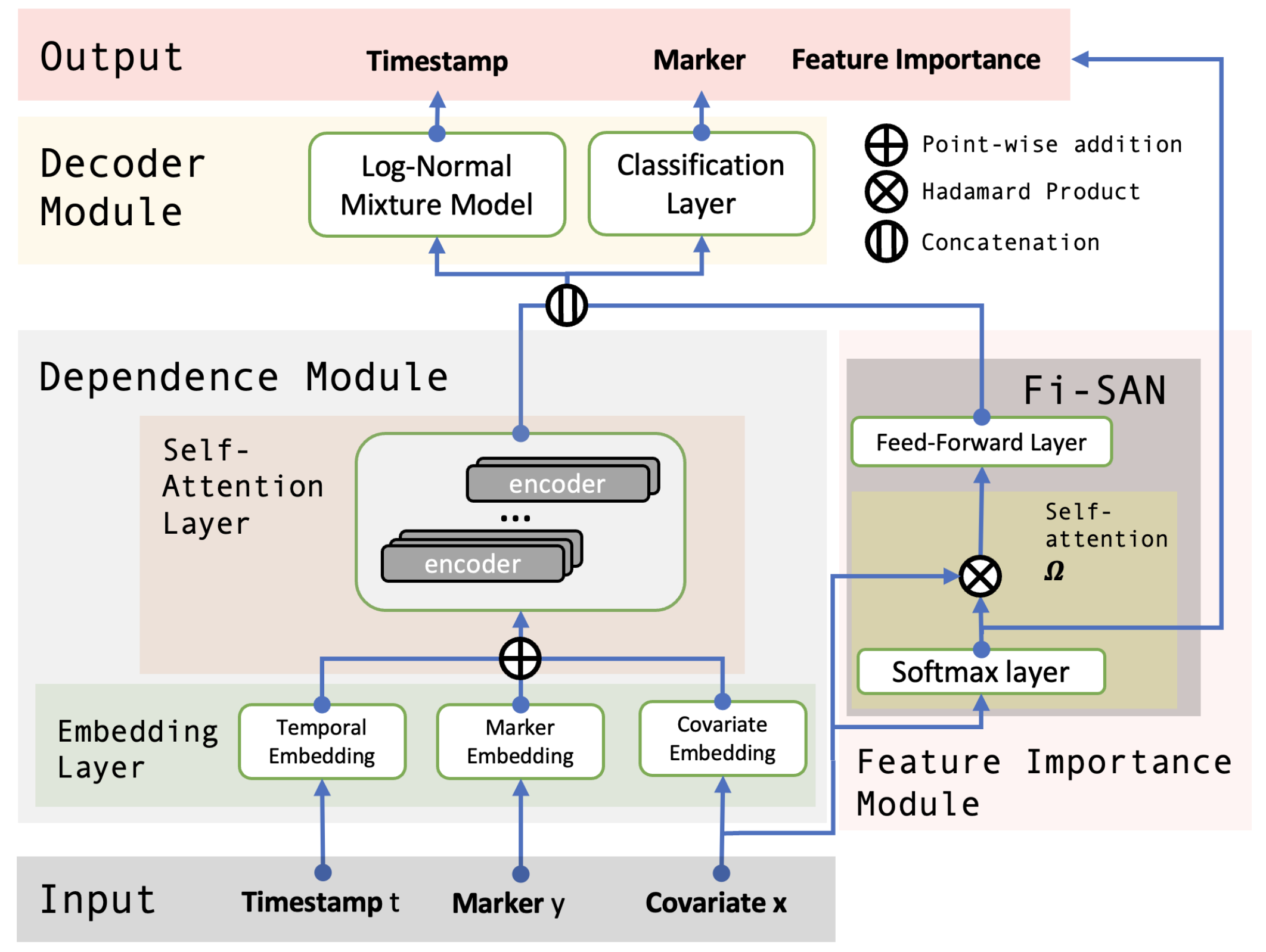}
    \caption{The model structure of TransFeat-TPP. TransFeat-TPP has three modules: the dependence module, the feature importance module and the decoder module. The dependence module extracts the dependencies among events to obtain the representation of the sequence; the feature importance module encodes the covariates and outputs their feature importance; the decoder module uses the learned representations from two previous modules to predict the next event's timestamp and type.}
    \label{fig: Model}
    \vspace{0.2in}
\end{figure}

In this work, we propose a Transformer-based covariate-TPP model with feature importance. This model is capable of incorporating covariate information to predict the occurrence time and type of future events, while simultaneously assessing the relative importance of various covariates. 
The model structure is illustrated in~\cref{fig: Model} where our model comprises three components: the dependence module, the feature importance module and the decoder module. In the following, we elaborate on the function and implementation of each module individually. 

\subsection{Problem Definition}

We present a concise definition of the problem we aim to address, along with an outline of the notations for the various variables employed throughout this work. 
Given a sequence of events $\mathcal{S}= \{(t_i,y_i,\mathbf{x}_i )\}_{i=1}^{L}$ where $L$ is the number of events and each event is represented by a triplet: $t_i \in \mathcal{R}^+$ is the timestamp that the $i$-th event occurs; $y_i$ represents the $i$-th event type which is a categorical variable; $\mathbf{x}_i \in \mathcal{R}^{F}$ is the covariate measured on the $i$-th event and $F$ is the covariate dimension. 
Given the historical information $\mathcal{H}_{t_n}=\{(t_i, y_i,\mathbf{x}_i)\}_{i=1}^n$, our objective is to predict the timestamp and type of the next event $(\hat{t}_{n+1}, \hat{y}_{n+1})$. 

\subsection{Dependence Module}
The dependence module of TransFeat-TPP is designed to encode the sequence and learn the dependencies among events. Drawing inspiration from \cite{zuo2020transformer}, we employ the Transformer architecture as the encoder. Specifically, we use the embedding layers to encode the triplets $(t_i, y_i, \mathbf{x}_i)$, and then adopt the self-attention layers to extract the dependencies among different events. 

\subsubsection{Embedding Layers}
Given a sequence $\mathcal{S}= \{(t_i,y_i,\mathbf{x}_i )\}_{i=1}^{L}$, we have three types of inputs: timestamp, event type and covariate. We need to encode three kinds of information to obtain an appropriate representation. 
We utilize temporal encoding for event timestamps, representing a given timestamp $t_i$ using an embedding vector:
\begin{equation*}
\left[\mathbf{z}\left(t_i\right)\right]_j=\left\{\begin{array}{l}
\cos \left(t_i / 10000^{\frac{j-1}{M}}\right) \text { if } j \text { is odd } \\
\sin \left(t_i / 10000^{\frac{j}{M}}\right) \text { if } j \text { is even }
\end{array}\right. \in \mathcal{R}^M,
\label{eq: temporal embedding per timestamps}
\end{equation*}
where $M$ is the dimension of temporal embedding, and $\left[\mathbf{z}\left(t_i\right)\right]_j$ denotes the $j$-th entry of $\mathbf{z}(t_i)$. Therefore, the collection of time embedding is: 
\begin{equation*}
    \mathbf{Z}=\left[\mathbf{z}(t_1), \mathbf{z}(t_2), \cdots, \mathbf{z}(t_L)\right]^\top \in \mathcal{R}^{L \times M}. 
\end{equation*}

In addition to temporal embedding, we employ a learnable matrix $\mathbf{U} \in \mathcal{R}^{M \times K}$ to encode event type, where the $k$-th column of $\mathbf{U}$ represents the $M$-dimensional embedding vector for event type $k$. For event type $y_i$, its embedding $\mathbf{e}(y_i)$ can be expressed as: 
\begin{equation*}
\label{eq:marker embedding per marker}
\mathbf{e}(y_i)=\mathbf{U} \mathbf{k}_i \in \mathcal{R}^M,
\end{equation*}
where $\mathbf{k}_i \in \mathcal{R}^K$ is the one-hot encoding of $y_i$. Therefore, the collection of type embedding is: 
\begin{equation*}
    \mathbf{E} = [ \mathbf{e}(y_1), \mathbf{e}(y_2), \cdots, \mathbf{e}(y_L)]^\top \in \mathcal{R}^{L \times M}. 
\end{equation*}

Analogous to event type embedding, we utilize a learnable matrix $\mathbf{W}\in\mathcal{R}^{M \times F}$ to encode covariates. For event covariate $\mathbf{x}_i$, its embedding $\mathbf{f}(\mathbf{x}_i)$ is expressed as: 
\begin{equation*}
\mathbf{f}(\mathbf{x}_i)=\mathbf{W}\mathbf{x}_i \in \mathcal{R}^M. 
\end{equation*}
Therefore, the collection of covariate embedding is: 
\begin{equation*}
    \mathbf{F}= [\mathbf{f}(\mathbf{x}_1), \mathbf{f}(\mathbf{x}_2), \cdots, \mathbf{f}(\mathbf{x}_L)]^\top \in \mathcal{R}^{L \times M}. 
\end{equation*}

Combining the previously mentioned three types of embeddings, we can obtain the final embedding of the sequence $\mathcal{S} = \{(t_i, y_i, \mathbf{x}_i)\}_{i=1}^L$:
\begin{equation}
\label{eq: overall embedding}
\mathbf{X} = \mathbf{Z} + \mathbf{E} + \mathbf{F} \in \mathcal{R}^{L \times M}, 
\end{equation}
where each row of $\mathbf{X}$ corresponds to the complete embedding of a single event in the sequence $\mathcal{S}$. 

\subsubsection{Self-attention Layers}
After the embedding layers, our focus shifts to learning the dependence among events. This is achieved by passing the representation $\mathbf{X}$ to the multi-head self-attention layers. The attention output $\mathbf{S}_h$ of the $h$-th head is formulated as: 
\begin{gather*}
\label{eq:transformer attention layer}
    \mathbf{S}_h  =  \text{softmax}\left(
        \frac{{\mathbf{Q}_h \mathbf{K}_h}^\top}{\sqrt{M_K}}\right)\mathbf{V}_h \in \mathcal{R}^{L \times M_V}, \\
    \mathbf{Q}_h = \mathbf{XW}^Q_h \in \mathcal{R}^{L \times M_K}, \\
    \mathbf{K}_h = \mathbf{XW}^K_h \in \mathcal{R}^{L \times M_K}, \\
    \mathbf{V}_h = \mathbf{XW}^V_h \in \mathcal{R}^{L \times M_V}, 
\end{gather*}
where $\mathbf{Q}_h$, $\mathbf{K}_h$, $\mathbf{V}_h$ are the query, key and value matrices of the $h$-th head obtained by different transformations from $\mathbf{X}$. $M_K$ is the dimension of $\mathbf{Q}_h$ and $\mathbf{K}_h$.
Matrices $\mathbf{W}^Q_h \in \mathcal{R}^{M\times M_K}$, $\mathbf{W}^K_h \in \mathcal{R}^{M\times M_K}$ and $\mathbf{W}^V_h \in \mathcal{R}^{M\times M_V}$ are the corresponding linear transformations of the $h$-th head. The final attention output is obtained by aggregating the output from $H$ heads: 
\begin{equation*}
    \mathbf{S} = \left[\mathbf{S}_1, \mathbf{S}_2, \cdots, \mathbf{S}_H\right]\mathbf{W}^O \in \mathcal{R}^{L \times M}, 
\end{equation*}
where $\mathbf{W}^O \in \mathcal{R}^{HM_V \times M}$ is a learnable matrix that is used to combine the outputs of each head. 
Finally, the attention output $\mathbf{S}$ is passed through a position-wise feedforward network that consists of two dense layers to produce the hidden state $\mathbf{H}_1$: 
\begin{equation}
\label{eq: transformer hidden}
    \mathbf{H}_1 = \text{ReLU}(\mathbf{SW}^{\text{FC}}_1 + \mathbf{b}_1)\mathbf{W}^{\text{FC}}_2 + \mathbf{b}_2 \in \mathcal{R}^{L \times M},
\end{equation}
where $\mathbf{W}^{\text{FC}}_1, \mathbf{W}^{\text{FC}}_2, \mathbf{b}_1, \mathbf{b}_2$ are the parameters of position-wise feedforward network. The hidden state $\mathbf{H}_1$ contains hidden representations of all events in the sequence, with each row corresponding to a specific event. 

\subsection{Feature Importance Module}

To determine the importance of covariates at the feature level, we utilize an additional Fi-SAN module. 
This module, designed for a specific classification task, learns to fit data and extract each feature's relevance to the outcome using self-attention. 
Given a covariate $\mathbf{x}_i$ as input, the Fi-SAN module produces two outputs: the auxiliary representation of the covariates and their feature importance. 

\subsubsection{Auxiliary Representation}

One of Fi-SAN's outputs is the auxiliary representation. To achieve this, $\mathbf{x}_i$ is firstly passed through a $\tilde{H}$-head self-attention layer $\mathbf{\Omega}$\footnote{The self-attention layer in Fi-SAN has a different structure with that in the dependence module. 
}:
\begin{equation}
    \mathbf{\Omega}(\mathbf{x}_i)=\frac{1}{\tilde{H}}\sum_{h=1}^{\tilde{H}}\left[ \mathbf{x}_i \otimes \text{softmax}(\mathbf{W}^{\text{FI}}_h \mathbf{x}_i + \mathbf{b}^{\text{FI}}_h)\right] \in \mathcal{R}^{F}, 
\label{eq: self-attention layers in Fi-SAN}
\end{equation}
where $\mathbf{W}^{\text{FI}}_h$ and $\textbf{b}^{\text{FI}}_h$ are the attention weights and bias of a softmax-activated layer of the $h$-th head. The softmax output can be interpreted as the importance score for each feature, which is then elementwise multiplied with $\mathbf{x}_i$ using the Hadamard product. The output of $\mathbf{\Omega}$ is the average over $\tilde{H}$ heads. 
The auxiliary representation is obtained from an additional dense layer on the top of $\mathbf{\Omega}$:
\begin{equation*}
\label{eq: Fi-SAN hidden}
    \begin{gathered}
    \mathbf{H}_2= \mathbf{\Omega}_\mathcal{S} \mathbf{W}^{\text{FC}}_3+\mathbf{b}^{\text{FC}}_3 \in \mathcal{R}^{L \times M},\\
    \mathbf{\Omega}_\mathcal{S}= [\mathbf{\Omega}(\mathbf{x}_1), \mathbf{\Omega}(\mathbf{x}_2), \cdots \mathbf{\Omega}(\mathbf{x}_L)]^{\top} \in \mathcal{R}^{L \times F},
    \end{gathered}
\end{equation*}
where $\mathbf{W}^{\text{FC}}_3$ and $\mathbf{b}^{\text{FC}}_3$ are the parameters of the dense layer, $\mathbf{\Omega}_\mathcal{S}$ is the result of passing all covariates $\{ \mathbf{x}_i \}_{i=1}^L$ through $\mathbf{\Omega}$. $\mathbf{H}_2$ is the auxiliary representation that is used for modelling event type. 

\subsubsection{Feature Importance}
Fi-SAN outputs the auxiliary representation while providing feature importance. Given the sequence $\mathcal{S}$, the final feature importance is the average of the feature importance across all points: 
\begin{gather*}
\label{eq: learned Fi on single covariates}
\text{FI}(\mathbf{x}_i) = \frac{1}{\tilde{H}}\sum_{h=1}^{\tilde{H}}\text{softmax}(\mathbf{W}^{\text{FI}}_h \mathbf{x}_i + \mathbf{b}^{\text{FI}}_h), \\
\text{FI}(\mathcal{S}) = \frac{1}{L} \sum_{i=1}^L \text{FI}(\mathbf{x}_i). 
\end{gather*}

\subsection{Decoder Module}
Our objective is to predict the time and type of the next event. To accomplish this, we design the decoder module, which is responsible for the next event's time and type prediction. 
To predict the timestamp of the next event, we utilize an intensity-free method proposed by \cite{shchur2019intensity}, which models the distribution of the next event's timestamp conditioned on the representation generated by the dependence module. To predict the type of the next event, we combine the representations from both dependence module and feature importance module, and use this combined representation to predict the next event type. 

\subsubsection{Timestamp Prediction}
\label{section: Timestamp Prediction}
To model the interevent time, denoted by $\tau$, which is the duration until the next event since the current event, we use a mixture of log-normal densities to represent the distribution of $\tau$, since $\tau$ must be positive (details provided in~\cref{app_log_normal_mixture}). The probability density function of a log-normal mixture distribution with $C$ components is: 
\begin{equation*}
\label{eq: log normal mixture distribution}
p(\tau|\mathbf{w}, \bm{\mu}, \mathbf{s}) = \sum_{c=1}^{C} w_c \frac{1}{\tau s_c \sqrt{2 \pi}}\exp\left(-\frac{(\log\tau - \mu_c)^2}{2s_c^2}\right), 
\end{equation*}
where $\mathbf{w}=[w_1,\ldots,w_C]$ is the mixture weight, $\bm{\mu}=[\mu_1,\ldots,\mu_C]$ and $\mathbf{s}=[s_1,\ldots,s_C]$ are the mixture mean and standard deviation. 
Let $\mathbf{h}_i$ be the representation for $i$-th event obtained by the dependence module (the $i$-th row of $\mathbf{H}_1$). We use $\mathbf{h}_i$ to define the parameters of the log-normal mixture distribution: 
\begin{gather*}
    \mathbf{w}_i=\text{softmax}(\mathbf{V}_w \mathbf{h}_i + \mathbf{b}_w), \ \ \ \ \bm{\mu}_i=\mathbf{V}_\mu \mathbf{h}_i + \mathbf{b}_\mu, \\
    \mathbf{s}_i=\exp(\mathbf{V}_s \mathbf{h}_i + \mathbf{b}_s), 
\end{gather*}
where $\mathbf{V}_w, \mathbf{V}_{\mu}, \mathbf{V}_s, \mathbf{b}_w, \mathbf{b}_\mu, \mathbf{b}_s$ are the learnable parameters. Consequently, the time loss is the negative log-likelihood given by: 
\begin{equation*}
    \mathcal{L}_1 = -\sum_{i=1}^L \log p(\tau_i|\mathbf{w}_i, \bm{\mu}_i, \mathbf{s}_i). 
\end{equation*}

\subsubsection{Type Prediction}
We concatenate the representation $\mathbf{H}_1$ obtained from the dependence module with the representation $\mathbf{H}_2$ obtained from the feature importance module, and feed it into a softmax-activated dense layer that maps from $\mathcal{R}^{2M}$ to $\mathcal{R}^{K}$, where $K$ is the number of event types: 
\begin{equation*}
    \mathbf{P} = \text{softmax}(\text{FC}([\mathbf{H}_1, \mathbf{H}_2])) \in \mathcal{R}^{L \times K}. 
\end{equation*}
The output of this layer provides the probability for each event type, so the cross entropy is used as the type loss: 
\begin{equation*}
    \mathcal{L}_2 = -\sum_i^L \log p_i(y_i), 
\end{equation*}
where $p_i(y_i)$ represents the probability of assigning $i$-th event to the correct label $y_i$. The total loss is the weighted sum of $\mathcal{L}_1$ and $\mathcal{L}_2$: 
\begin{equation*}
\label{Total loss}
    \mathcal{L} = \alpha_1 \mathcal{L}_1 + \alpha_2 \mathcal{L}_2, 
\end{equation*}
where $\alpha_1$ and $\alpha_2$ are the parameters to balance the loss. 
We utilize the trick introduced in \cite{kendall2018multi} to automatically adjust the weights based on the current magnitude of each loss component.

\section{Experiments}
We evaluate TransFeat-TPP on both synthetic and real datasets. We compare TransFeat-TPP to several deep TPP models and find that it is more flexible and interpretable. This is because TransFeat-TPP enables more expressive incorporation of covariates. At the same time, TransFeat-TPP is capable of effectively learning consistent feature importance of covariates for event type prediction, thus demonstrating its potential as an interpretable deep covariate-TPP. 

\subsection{Baseline}
\label{section: baselines}
We compare our model against existing deep TPP models: recurrent marked temporal point process (\textbf{RMTPP})~\cite{du2016recurrent}, recurrent neural network point process (\textbf{RNNPP})~\cite{xiao2017intensity}, Transformer Hawkes process (\textbf{THP})~\cite{zuo2020transformer}.


RMTPP uses a recurrent neural network to learn the conditional intensity function by encoding event time and type, but it does not include covariates. To enable fair comparison, we propose \textbf{RMTPP+}, which incorporates covariate information. RNNPP models the temporal point process by incorporating both event and covariate information but offers limited interpretability regarding covariate relevance. THP employs the Transformer for enhanced computational efficiency, encoding only event time and type. We propose \textbf{THP+}, a modified version that includes covariate information. Details about RMTPP+ and THP+ are in \cref{app_modified_model}.

\subsection{Evaluation Metric}
\label{section: Evaluation Metric}
We evaluate the performance using a variety of widely-accepted metrics. (1) \textbf{log-likelihood}: A typical metric to evaluate the TPP model is the log-likelihood on test sets. However, RNNPP does not model the conditional intensity function and therefore it does not produce log-likelihood. As a result, we only compare to RMTPP and THP in terms of log-likelihood. (2) \textbf{Time Prediction}: For event time prediction, we employ the root mean square error (RMSE) as a metric to assess the discrepancy between the predicted and actual event times. (3) \textbf{Type Prediction}: For event type prediction, we use accuracy and F1 score as evaluation metrics. It is worth noting that in a multi-class classification problem, the F1 score is calculated as a weighted average across all classes. 

\begin{table*}[tb]
\centering
\caption{Experimental results from six different models on two synthetic datasets. The enhanced performance of the modified models, RMTPP+ and THP+, highlights the benefits of incorporating covariates. The superiority of TransFeat-TPP demonstrates the improvement in model expressiveness and flexibility. Standard deviation in brackets. }
\label{table: synthetic Experiment Result}
\begin{sc}
\scalebox{0.8}{
\begin{tabular}{ccccc|ccccc} 
\toprule
\multirow{2}{*}{Model} & \multicolumn{4}{c}{Inhomogeneous} & \multicolumn{4}{c}{Hawkes} \\
\cmidrule{2-9}
    & LOG-LL & RMSE & ACC & F1 & LOG-LL & RMSE & ACC & F1\\ 
\midrule
\multirow{2}{*}{RMTPP}  & -3.07 & 2.77 & 58.11\% & 0.3844 & -4.82 & 1.90 & 64.58\%  & 0.5165\\
& (0.092) & (0.075)  & (0.010)  & (0.009)  & (0.091) & (0.113)  & (0.042)  & (0.032) \\

\multirow{2}{*}{RMTPP+} & -2.89 & 2.80 & 88.30\% & 0.8950 &-3.98 & 1.75 & 84.04\% & 0.8109 \\
& (0.079) & (0.160)  & (0.003)  & (0.010) & (0.073) & (0.094)  & (0.033)  & (0.015)  \\

\multirow{2}{*}{RNNPP} & --- & 3.26 & 85.63\% & 0.8105 & --- & 1.89 & 85.33\% & 0.8311\\
& --- & (0.093)  & (0.033)  & (0.013) & --- & (0.077)  & (0.041) & (0.033)\\

\multirow{2}{*}{THP} & -4.77 & 3.53 & 63.48\% & 0.4188 & -3.15 & 1.85 & 71.39\% & 0.5415\\
& (0.043) & (0.054)  & (0.098) & (0.077)  & (0.092) & (0.071)  & (0.043) & (0.069)\\

\multirow{2}{*}{THP+} & -4.65 & 2.87 & 89.01\% & 0.8901 & -3.01 & 1.84 & 86.51\% & 0.8663\\
& (0.019) & (0.054)  & (0.033) & (0.078) & (0.037) & (0.043)  & (0.013) & (0.025)\\

\multirow{2}{*}{\textbf{TransFeat}} & \textbf{-1.10} & \textbf{2.41} &\textbf{91.81\%} & \textbf{0.9132} 
& \textbf{-2.37}& \textbf{1.58} & \textbf{89.06\%} & \textbf{0.8725}\\
& (0.053) & (0.039)  & (0.023) & (0.018) & (0.037) & (0.050)  & (0.015) & (0.025)\\
\bottomrule
\end{tabular}}
\end{sc}
\end{table*}

\begin{figure}[t]
    \begin{center}
    \begin{minipage}{0.48\linewidth}
    \includegraphics[width=\columnwidth]{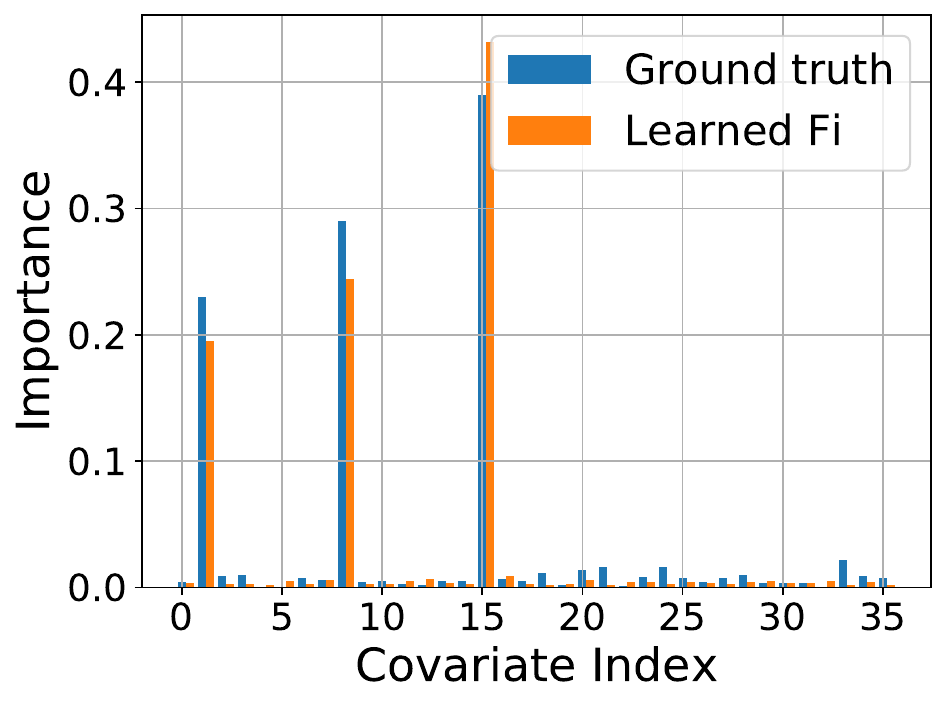}
    \subcaption[]{Hawkes-1}
    \label{fig: Fi study:a}
    \end{minipage}%
    \begin{minipage}{0.48\linewidth}
    \includegraphics[width=\columnwidth]{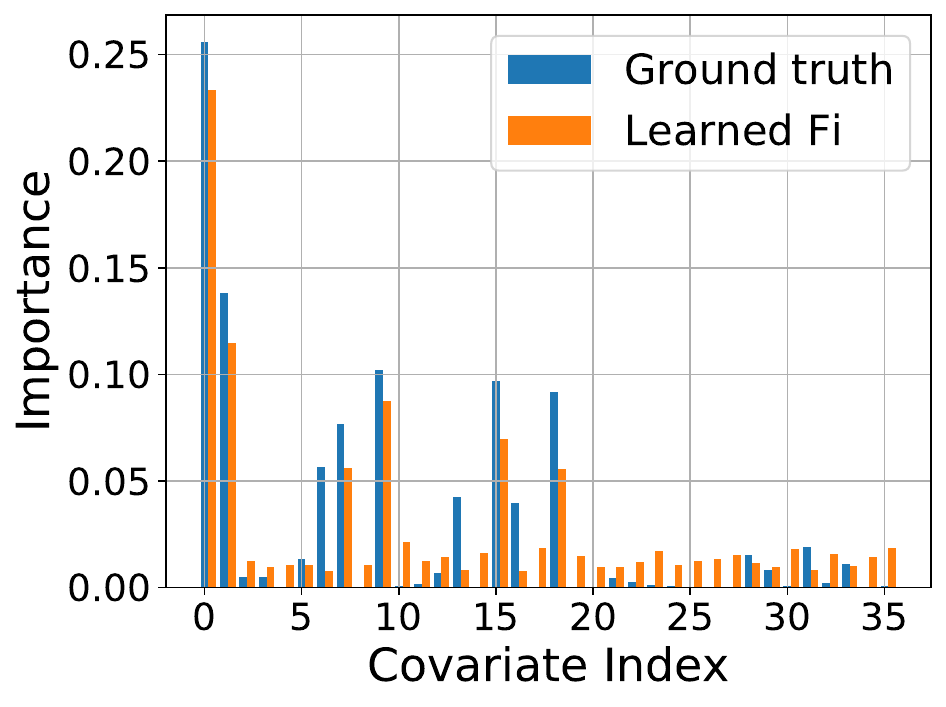}
    \subcaption[]{Hawkes-2}
    \label{fig: Fi study:b}
    \end{minipage}\\
    \vspace{0.1in}
    \begin{minipage}{0.48\linewidth}
    \includegraphics[width=\columnwidth]{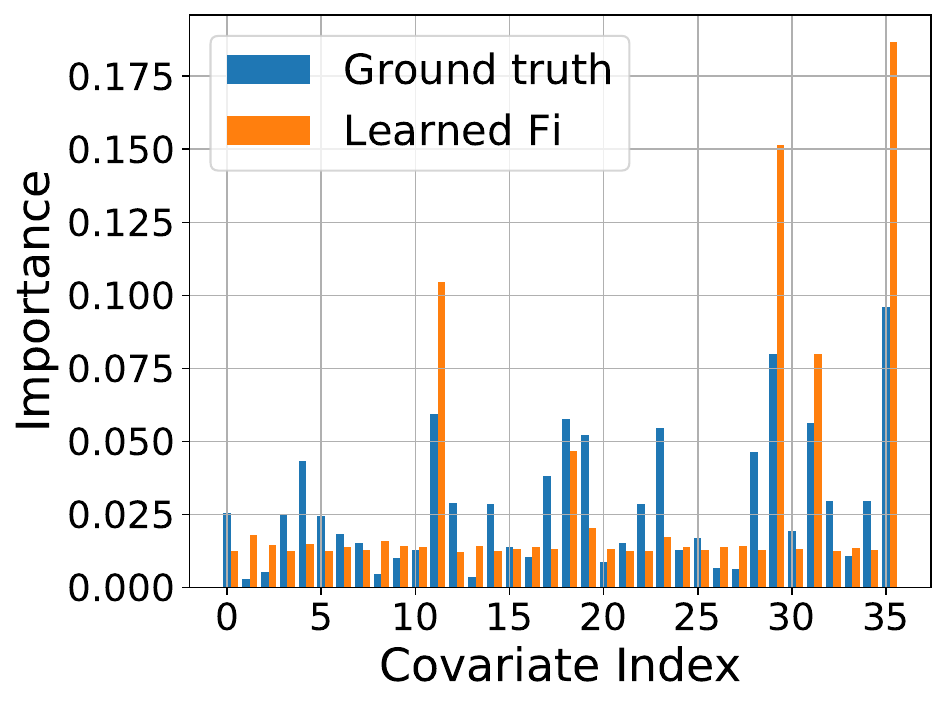}
    \subcaption[]{Hawkes-3}
    \label{fig: Fi study:c}
    \end{minipage}%
    \hspace{0.05in}
    \begin{minipage}{0.48\linewidth}
    \includegraphics[width=\columnwidth]{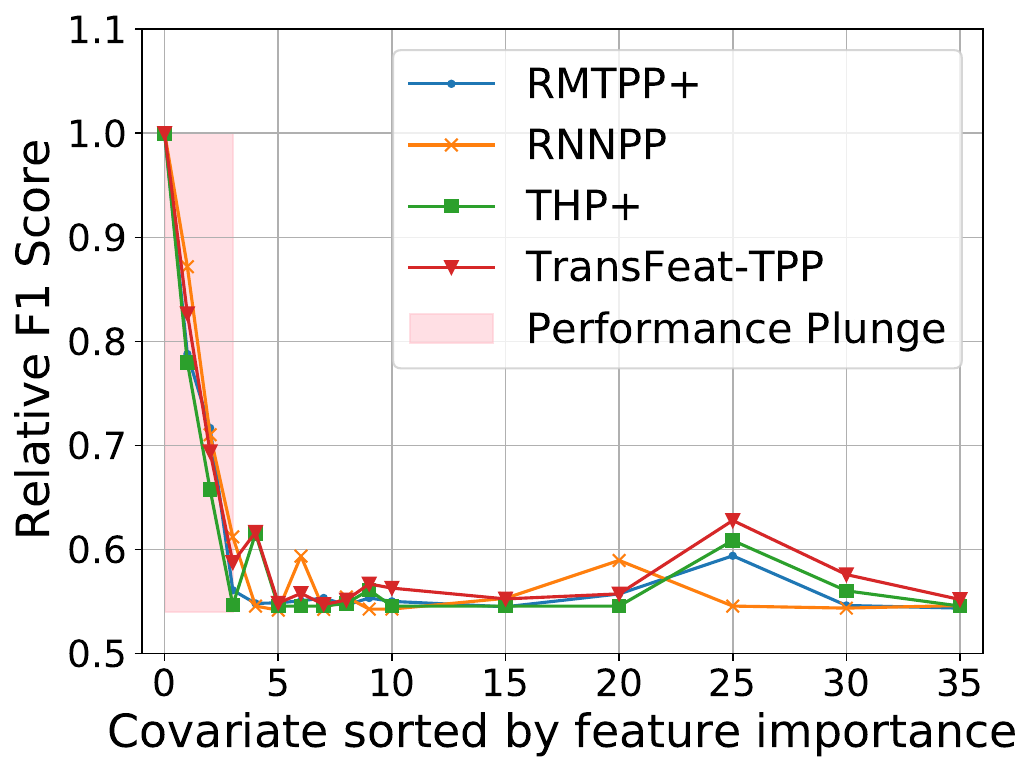}
    \subcaption[]{Hawkes-1}
    \label{fig: Fi study:d}
    \end{minipage}\\
    \vspace{0.1in}
    \begin{minipage}{0.48\linewidth}
    \includegraphics[width=\columnwidth]{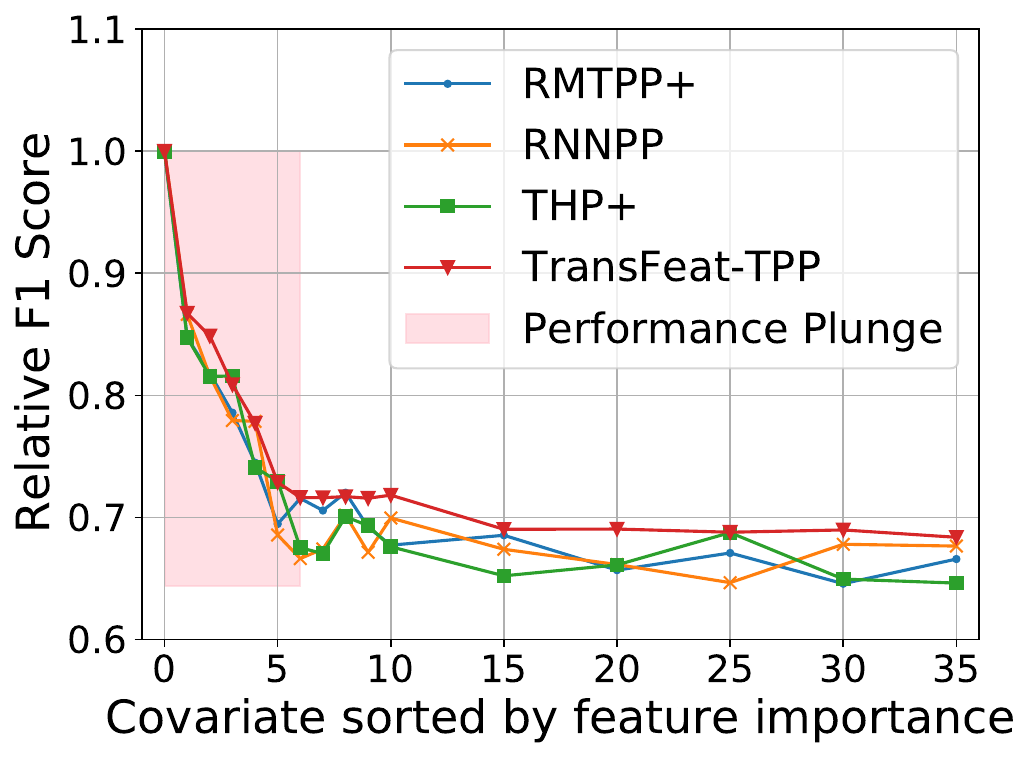}
    \subcaption[]{Hawkes-2}
    \label{fig: Fi study:e}
    \end{minipage}%
    \begin{minipage}{0.48\linewidth}
    \includegraphics[width=\columnwidth]{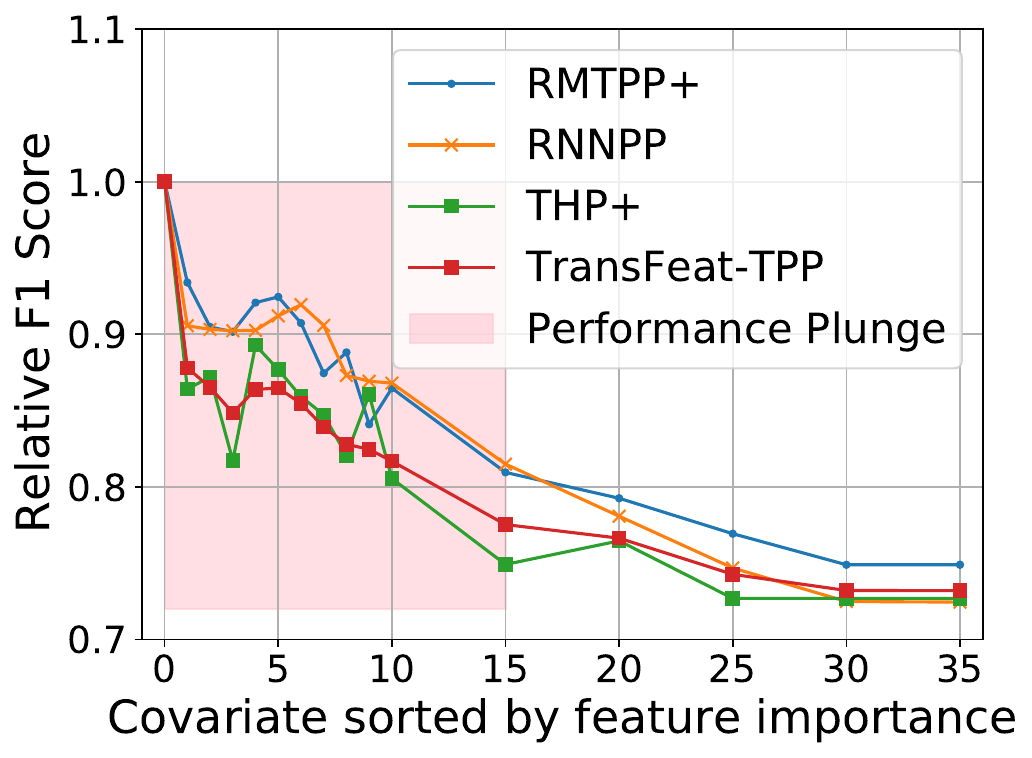}
    \subcaption[]{Hawkes-3}
    \label{fig: Fi study:f}
    \end{minipage}
    \end{center}
    \caption{The feature importance comparison and ablation study. (a)(b)(c): We compare the learned feature importance with the ground truth, verifying that TransFeat-TPP can differentiate between important and irrelevant covaraites. (d)(e)(f): We show the feature ablation study to indicate the consistency of the learned feature ranking. The x-axis represents the covariates sorted from most to least important. When more important covariates are removed, the models experience a significant performance drop. As less impactful covariates are removed, the performance decline becomes smoother.}
    \label{fig: Fi study}
    \vspace{0.2in}
\end{figure}

\subsection{Synthetic Data}
\label{section: Synthetic Dataset}
In this section, we perform experiments on the synthetic covariate-TPP data, designed to resemble TPPs influenced by covariates. 
We verify that on one hand, by incorporating covariate information, our model can achieve better predictive performance; on the other hand, our model can uncover the latent feature importance of covariates. 

\subsubsection{Data Simulation}
\label{section: synthetic setup}
For the synthetic data, we assume covariates can impact both the arrival rate and type of future events. This is implemented by incorporating covariates into the conditional intensity function and considering covariates when assigning future events' types. 
Given historical events: $\{(t_i, y_i, \mathbf{x}_i)\}_{i=1}^n$, we assume the conditional intensity function can be written as: $\lambda(t)=f(\mathbf{x}_{n},\mathcal{H}_{t})$ where $\mathcal{H}_{t}$ is the historical information before $t$ and $\mathbf{x}_{n}$ is the latest covariate before $t$. We use the Ogata thinning algorithm \cite{ogata1981lewis} to simulate the arrival time $t_{n+1}$. 
After that, we assume the event type $y_{n+1}$ is binary and simulate it according to a Bernoulli distribution whose parameter $p$ depends on the latest covariate $\mathbf{x}_{n}$ and the historical information $\mathcal{H}_{t}$: $p=g(\mathbf{x}_{n},\mathcal{H}_{t})$. 
It is worth noting that different point process models correspond to different functional forms of $f$ and $g$. 

\subsubsection{Prediction Task}
This section aims to demonstrate the improved predictive performance of our model through the incorporation of covariate information in an expressive way. To this end, we simulate two different types of covariate-TPP datasets, covariate inhomogeneous Poisson process and covariate Hawkes process, to validate the advantage of our model. The simulation details and statistics of the synthetic datasets are provided in~\cref{Appendix: Synthetic Data}. We can observe that RMTPP+ and THP+ outperform their respective original models, indicating that the incorporation of covariates is beneficial for event prediction. Moreover, our proposed model TransFeat-TPP surpasses other baselines in all cases, which can be attributed to the incorporation of Fi-SAN. Fi-SAN concentrates on the covariates and delivers an auxiliary representation, enhancing the model's expressiveness and flexibility in handling covariates.

\subsubsection{Feature Importance Task} 

Another advantage of our model is its ability to reveal the latent feature importance of covariates, thereby enhancing the model's interpretability. 
To achieve this, we validate our model on three Hawkes processes, labeled as Hawkes-1, 2 and 3, each with a unique feature importance configuration, corresponding to three levels of difficulty. Note that the Hawkes-1 is the ``Hawkes'' synthetic dataset presented in~\cref{table: synthetic Experiment Result}, while Hawkes-2 and 3 have more complex ground truth feature importance, making it more challenging for the model to distinguish between them. 
We present two categories of results: a comparison of feature importance and an ablation study. For the feature importance comparison, \cref{fig: Fi study:a,fig: Fi study:b,fig: Fi study:c} demonstrate our learned feature importance in comparison to the ground truth. For Hawkes-1 where most covariates are irrelevant, TransFeat-TPP can readily capture the most important ones. As more covariates influence the event dynamics (Hawkes-2 and 3), our model remains capable of distinguishing the most important ones. The experimental results show that TransFeat-TPP can effectively identify the most important covariates while assigning lower importance to irrelevant ones. In the feature ablation study, we examine the impact of removing covariates according to the learned importance ranking. The results in \cref{fig: Fi study:d,fig: Fi study:e,fig: Fi study:f} indicate that the model's performance experiences a significant decline when the top important covariates are removed, followed by a more gradual decrease when irrelevant ones are removed. This confirms the consistency of the learned ranking of feature importance. 

\begin{table*}[tb]
\centering
\caption{Experimental results on two real datasets demonstrate the superior performance of modified models, RMTPP+ and THP+, compared to the original models. These findings provide compelling evidence that incorporating covariates enhances the predictive capabilities of the models. Our proposed TransFeat-TPP achieves the best performance in the majority of cases, highlighting the effectiveness of the Fi-SAN module in introducing greater flexibility and expressiveness. Note that the ``NO Fi-SAN'' indicates the TransFeat-TPP model without the Fi-SAN module, which is used for ablation study in \cref{subsection: ablation study on Fi-SAN module}. Standard deviation in brackets. For the PM2.5 dataset, we present results for Beijing and Shenyang here, while additional experimental results for other cities can be found in \cref{table: Extra Experiment Result} in the Appendix.} 
\label{table: real data Experiment Result}
\begin{sc}
\scalebox{0.8}{
\begin{tabular}{ccccc|cccc|cccc} 
\toprule
\multirow{2}{*}{Model} & \multicolumn{4}{c}{PM2.5-Bejing} & \multicolumn{4}{c}{PM2.5-Shenyang} & \multicolumn{4}{c}{Car-accident-London}\\
\cmidrule{2-13}
    & LOG-LL & RMSE & ACC & F1 & LOG-LL & RMSE & ACC & F1 & LOG-LL & RMSE & ACC & F1\\ 
\midrule
\multirow{2}{*}{RMTPP} & -2.03 & 0.2876 & 45.11\% & 0.4210 & -1.84 & 0.2564 & 44.34\% & 0.4187 & -2.89 & 1.690 & 59.40\% & 0.5447\\
 & (0.012) & (0.033) & (0.017) & (0.011) & (0.020) & (0.056) & (0.011) & (0.009) & (0.095) & (0.101) & (0.054)& (0.031) \\

\multirow{2}{*}{RMTPP+} & -1.64 & 0.2213 & 47.56\% & 0.4653 & -1.66 & 0.2255 & 49.34\% & 0.4887 & -2.78 & 1.271 & 65.03\% & 0.5689\\
 & (0.035) & (0.029) & (0.035) & (0.031) & (0.020) & (0.025) & (0.018) & (0.016) & (0.036) & (0.089) & (0.082)& (0.096)\\
 
\multirow{2}{*}{RNNPP} & --- & 0.2319 & 49.56\% & 0.4811 & --- & 0.34 & 45.34\% & 0.4377 & --- & 0.9181 & 63.06\% & 0.5333\\
 & --- & (0.010) & (0.017) & (0.036) & --- & (0.021) & (0.025) & (0.019) & --- & (0.014) & (0.025) & (0.013) \\

\multirow{2}{*}{THP} & -1.97 & 0.5094 & 44.28\% & 0.3572 & -2.24 & 0.3125 & 41.28\% & 0.3272 & -2.01 & 0.8994 & 58.12\% & 0.5289\ \\

 & (0.077) & (0.028) & (0.033) & (0.036) & (0.023) & (0.021) & (0.025) & (0.019) & (0.049) & (0.030) & (0.061) & (0.033) \\

\multirow{2}{*}{THP+} & -1.86 & 0.4930 & 50.89\% & 0.4883 & -2.16 & 0.3881 & 49.53\% & 0.4830  & -1.98 & 0.8710 & 64.13\% & 0.5579 \\
 & (0.006) & (0.091) & (0.005) & (0.013) & (0.009) & (0.083) & (0.002) & (0.022) & (0.031) & (0.082) & (0.003) & (0.029) \\

\multirow{2}{*}{\textbf{TransFeat}} & \textbf{-1.30} & \textbf{0.1619} &\textbf{51.70}\% & \textbf{0.4891} 
& \textbf{-1.31}& \textbf{0.1993} & \textbf{51.24}\% & \textbf{0.5338}
& \textbf{-1.60}& \textbf{0.6420} & \textbf{66.33} \% & \textbf{0.6153}\\

& (0.051) & (0.070) & (0.089) & (0.066)
& (0.016) & (0.051) & (0.071) & (0.035)  & (0.041) & (0.009) & (0.046) & (0.015) \\

\midrule
\multirow{2}{*}{NO Fi-SAN} & -1.32 & 0.1703 &49.07\% & 0.4779 
& -1.34 & 0.2384 & 49.72\% & 0.4818
& \textbf{-1.60} & 0.6473 & 63.52\%& 0.6014 \\

& (0.010) & (0.084) & (0.068) & (0.054)
& (0.027) & (0.083) & (0.067) & (0.044)  & (0.021) & (0.011) & (0.013) & (0.028) \\
\bottomrule
\end{tabular}}
\end{sc}
\end{table*}

\begin{figure*}[tb]
    \begin{center}
    \adjustbox{valign=b}{
    \begin{minipage}{0.24\textwidth}
    \includegraphics[width=\columnwidth]{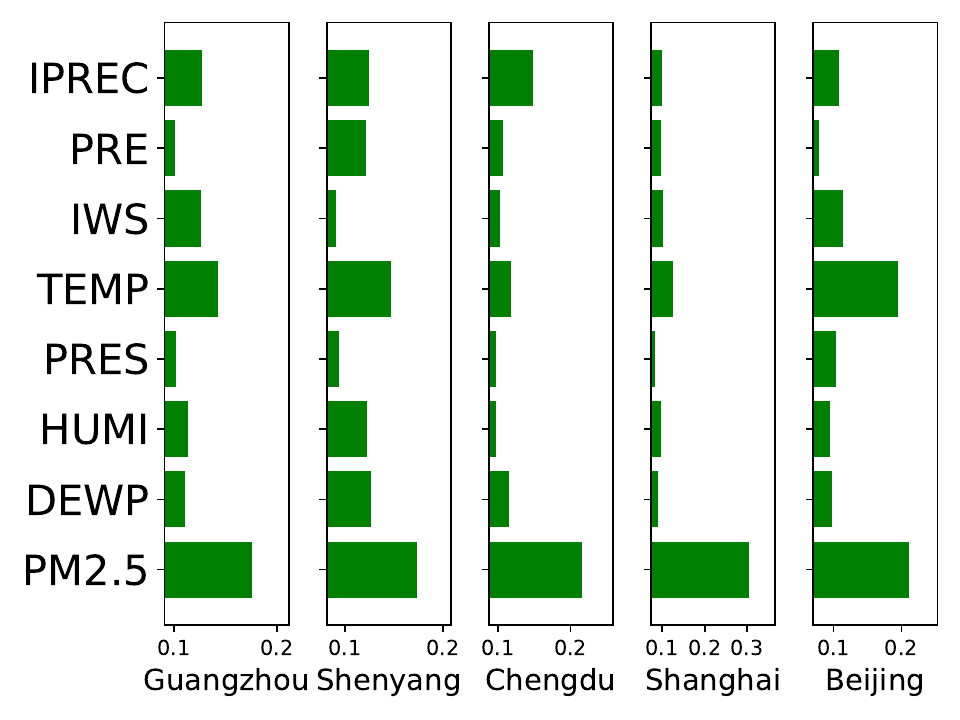}
    \subcaption[]{Learned feature importance for PM2.5 Alert Dataset}
    \label{fig: PM2.5 Fi study:a}
    \end{minipage}}
    \adjustbox{valign=b}{
    \begin{minipage}{0.24\textwidth}
    \includegraphics[width=\columnwidth]{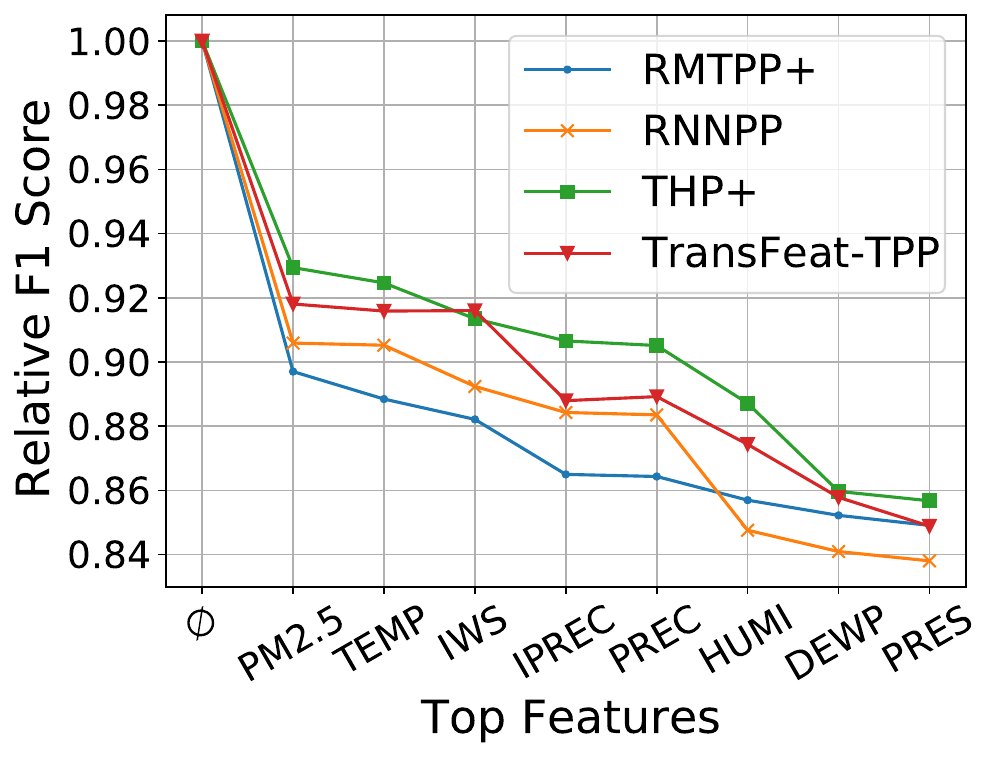}
    \subcaption[]{Feature ablation study for PM2.5 Alert Dataset}
    \label{fig: PM2.5 Fi study:b}
    \end{minipage}}\scalebox{0.96}{
    \adjustbox{valign=b}{
    \begin{minipage}{0.24\textwidth}
    \includegraphics[width=\columnwidth]{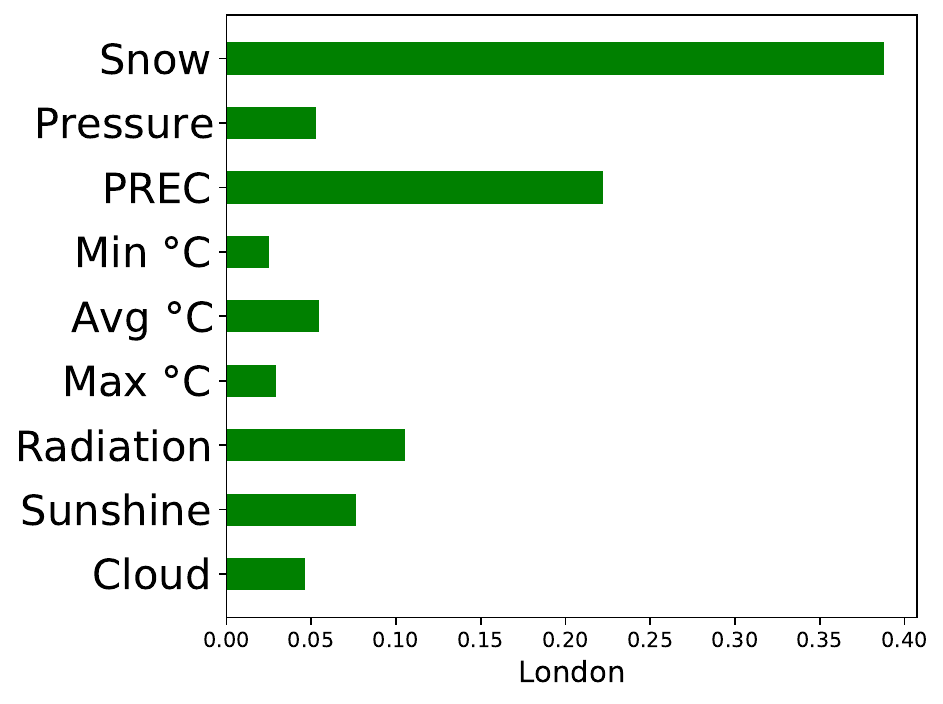}
    \subcaption[]{Learned feature importance for London Car Accident Dataset}
    \label{fig: london Fi study:a}
    \end{minipage}}}
    \adjustbox{valign=b}{
    \begin{minipage}{0.24\textwidth}
    \includegraphics[width=\columnwidth]{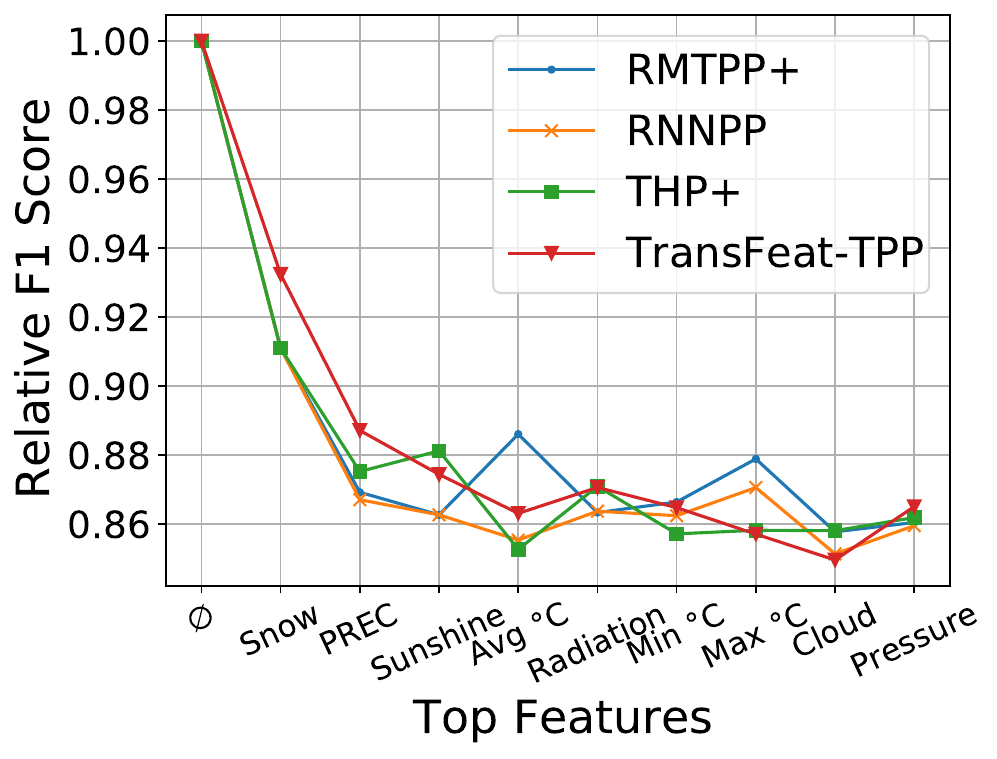}
    \subcaption[]{Feature ablation study for London Car Accident Dataset}
    \label{fig: london Fi study:b}
    \end{minipage}}
    \end{center}
    \caption{(a)(c): The learned feature importance for the PM2.5 alert dataset and the London car accident dataset, respectively. For the PM2.5 dataset, the two most important features are the current PM2.5 concentration and temperature, while for the car accident dataset, the two most critical environmental factors impacting accident severity are snow depth and precipitation. (b)(d): The feature ablation study for the PM2.5 alert dataset and the London car accident dataset. The x-axis represents the covariates sorted from most to least important. When the most important covariate (current PM2.5 concentration for the PM2.5 alert dataset and snow depth for the London car accident dataset) is removed, the models experience a significant performance drop. As less impactful covariates are removed, the performance decline becomes smoother.}
    \label{fig: PM2.5 Fi study}
    \vspace{0.2in}
\end{figure*}

\subsection{Real Data}
In this section, we verify our model on two real datasets: PM2.5 alert dataset and London car accident dataset. 

\subsubsection{PM2.5 Alert Dataset}

PM2.5, a significant contributor to air pollution, triggers air pollutant alerts based on specific concentration thresholds. These alerts, characterized by timestamps and severity types, are considered noteworthy events. Besides, we also consider the hourly meteorological records as relevant covariates, which contain extra information such as temperature, humidity, precipitation, etc. More details can be found in~\cref{Appendix: Real-Data PM2.5}. The dataset\footnote{\url{https://archive.ics.uci.edu/ml/datasets/PM2.5+Data+of+Five+Chinese+Cities}} was collected from five Chinese cities (Beijing, Shenyang, Shanghai, Chengdu, and Guangzhou) between 2010 and 2015. For each city, separate models are trained using data from 2010 to 2013 for training, 2014 for validation, and 2015 for testing. 

\subsubsection{London Car Accident Dataset}

Car accidents are a prevalent phenomenon in the real world and have attracted considerable research attention~\cite{li2018traffic, bergel2013explaining}. 
We analyze a car accident dataset\footnote{\url{https://www.kaggle.com/datasets/silicon99/dft-accident-data}} containing all car accidents in London from 2005 to 2015. Additionally, relevant meteorological data\footnote{\url{https://www.kaggle.com/datasets/emmanuelfwerr/london-weather-data}} is considered as covariates for our analysis. Further information on both real datasets can be found in~\cref{Appendix: Real Data}. 


\subsubsection{Prediction Task}
We evaluate our model's predictive performance on the two mentioned real-world datasets. 
We reach a consistent conclusion on the real dataset that aligns with our findings on the synthetic dataset. 
Results shown in~\cref{table: real data Experiment Result} indicate that the modified versions RMTPP+ and THP+ outperform their original counterparts in most cases.
This indicates that incorporating meteorological data as covariates potentially improves the model's predictive capabilities for air pollution alerts and car accident predictions. 
Furthermore, our proposed TransFeat-TPP outperforms other baselines or achieves competitive results, which validates that the incorporation of Fi-SAN enables our model to be more expressive and flexible.

\subsubsection{Feature Importance Task}
We analyze the feature importance of covariates in relation to the future PM2.5 alert type. The results in \cref{fig: PM2.5 Fi study:a,fig: PM2.5 Fi study:b} provide practical insights into which covariate has a significant impact on the future PM2.5 alert type. 
The comprehensive analysis leads us to a general conclusion that, consistently across various cities, the current concentration of PM2.5 plays a pivotal role in determining the type of future PM2.5 air quality alerts. This finding is both logical and reasonable. This is because the occurrence of an air pollution alert implies that the concentration of PM2.5 has transitioned from the current stage to a subsequent stage. 
Meanwhile, the temperature is the second most important factor in most cities, which implies the seasonal variation pattern of PM2.5. This pattern has been studied and verified by numerous works~\cite{huang2018spatial, kong2020investigating, zheng2005seasonal}. Specifically, \cite{kong2020investigating} proposed that significant seasonal variations are observed in Beijing, where the highest PM2.5 concentrations are typically observed in the winter, and the lowest are generally found in the summer. 
Similarly, we investigate the potential correlation between weather conditions and the severity levels of car accidents occurring during winter in London. In \cref{fig: london Fi study:a,fig: london Fi study:b}, we can observe two significant covariates: snow depth and precipitation, which have a notable impact on accident severity. This finding is intuitive and aligns with common sense since both snowfall and rainfall can result in slippery road conditions and reduced visibility, potentially leading to more severe car accidents. In fact, this phenomenon has been studied and confirmed by numerous previous works~\cite{fountas2020joint,liu2020analysis}. 

\subsection{Ablation Study on Fi-SAN Module}
\label{subsection: ablation study on Fi-SAN module}

We conduct an extra ablation study on the Fi-SAN module. Specifically, we remove the Fi-SAN module to create an ablation model named as ``NO Fi-SAN''. We compare its performance with other baseline models in \cref{table: real data Experiment Result}. 
The results indicate that NO Fi-SAN demonstrates similar log-likelihood and RMSE performance but inferior classification performance (ACC, F1) compared to the original TransFeat-TPP. This suggests that Fi-SAN contributes to the classification performance more by introducing additional parameters into the model. 
In contrast, NO Fi-SAN surpasses all baselines except TransFeat-TPP in terms of log-likelihood and RMSE, which is attributed to the use of mixture log-normal modeling, a method previously validated by \cite{shchur2019intensity}. 
Therefore, we conclude that our TransFeat-TPP outperforms other models in predictive performance due to three key factors: 
(1) the incorporation of covariates (more relevant information); (2) the leveraging of the Fi-SAN module (improved event type prediction); (3) the utilization of mixture log-normal modeling (improved event time prediction). 

\subsection{Hyperparameter Analysis}

\label{hyperparameter analysis}


Our model's configuration primarily encompasses several dimensions: the encoding dimension of temporal and type embeddings, denoted as $M$; the dimension of query and key embeddings, denoted as $M_Q$ and $M_K$ where $M_Q=M_K$; the dimension of the value embeddings, denoted as $M_V$; the dimension of auxiliary representation, denoted as $F$; the number of attention heads in the dependence module and Fi-SAN, denoted as $H$ and $\tilde{H}$. We test the sensitivity of our model to these hyperparameters by using various configurations. Specifically, we set $M=M_Q=M_K$, $H=\tilde{H}$, and conduct the hyperparameter analysis on one synthetic dataset and one public dataset: the Hawkes-1 dataset and the PM2.5 Beijing dataset. The results are shown in \cref{table: Hyper-Parameter Configurations} which indicates that our model is not significantly affected by the hyperparameter variation. Besides, it can achieve reasonably good performance even with fewer parameters. 

\begin{table}[t]
    \centering
    \caption{We experiment with different hyperparameter configurations on one toy dataset (Hawkes-1) and one public dataset: PM2.5-Beijing. Note that we set $M=M_Q=M_K$ and $H=\tilde{H}$. The results indicate that our model is robust to these hyperparameters.}
    \begin{sc}
    \scalebox{0.78}{
    \begin{tabular}{ccccc}
        \toprule
        \multirow{2}{*}{Config} & \multicolumn{2}{c}{PM2.5-BEIJING} & \multicolumn{2}{c}{HAWKES-1}\\
         & LOG-LL & ACC & LOG-LL & ACC\\
        \midrule
        $M=32$, $M_V=16$, $H=1$, $F=32$ & -1.58 & 47.19\% & -2.48& 88.00\%\\
        $M=64$, $M_V=32$, $H=2$, $F=64$ &-1.30  & 51.70\% & -2.37 & 89.06\%\\
        $M=128$, $M_V=64$, $H=2$, $F=128$ & -1.29 & 52.03\% & -2.39& 88.12\%\\
        $M=128$, $M_V=64$, $H=4$, $F=128$ & -1.30 & 51.11\% & -2.40 & 87.84\%\\
        \bottomrule    \end{tabular}}
    \end{sc}
    \label{table: Hyper-Parameter Configurations}
\end{table}

\section{Conclusion and Discussion}
We explore interpretability in covariate-TPPs and propose a novel TransFeat-TPP, a model that incorporates covariates in event modeling and extracts insights into feature importance. 
Our experiments show that incorporating covariates enhances event modelling, with our model outperforming other baselines, and the modified versions (RMTPP+, THP+) outperforming their original versions (RMTPP, THP). 
The ablation study supports the consistency of our learned feature importance ranking, highlighting the significant contribution of top features and the minimal impact of less important ones. 
Our proposed TransFeat-TPP effectively manages covariates and offers deeper insights than other baselines. This work is a significant advancement in exploring interpretability in deep TPPs. 
For future investigations, we aim to delve further into the interpretability of deep temporal point process models. Specifically, we plan to explore additional aspects, such as identifying which covariates influence specific types of events and understanding the nature of their effects, for insance, stimulation or inhibition. By addressing these aspects, we aim to gain a more comprehensive understanding of the role of covariates in temporal point process modelling. 

\section*{Acknowledgments}
This work was supported by NSFC Project (No. 62106121), the MOE Project of Key Research Institute of Humanities and Social Sciences (22JJD110001), and the Public Computing Cloud, Renmin University of China.

\bibliography{mybibfile}

\clearpage
\newpage
\appendix
\section{Modified Model: RMTPP+ and THP+}
\label{app_modified_model}
In this section, we introduce the details of RMTPP+ and THP+. Given the last hidden state $h_{i-1}$, current event time $t_i$, current event type $y_i$, and covariate $\mathbf{x}_i$, the original RMTPP takes the concatenation of time and event type embeddings as the RNN's input, while RMTPP+ takes the input concatenated with an additional covariate to generate the hidden state: 
\begin{equation*}
    \begin{gathered}
        \text{RMTPP}: \mathbf{h}_{i} = \text{RNN}([t_i, y_i], \mathbf{h}_{i-1}), \\
        \text{RMTPP+}: \mathbf{h}_{i} = \text{RNN}([t_i, y_i, \mathbf{x}_i ], \mathbf{h}_{i-1}). 
    \end{gathered}
\end{equation*}

Similarly, the original THP takes temporal embedding $\mathbf{Z}$ and type embedding $\mathbf{E}$ as input only, while THP+ incorporates the additional covariate embedding $\mathbf{F}$ as input: 
\begin{equation*}
    \begin{gathered}
        \text{THP}: \mathbf{X} = \mathbf{Z} + \mathbf{E}, \\
        \text{THP+}: \mathbf{X} = \mathbf{Z} + \mathbf{E} + \mathbf{F}, 
    \end{gathered}
\end{equation*}
where $\mathbf{Z}$, $\mathbf{E}$, and $\mathbf{F}$ refer to the temporal, type, and covariate embeddings as described in~\cref{eq: overall embedding}.

\section{Synthetic Data}
\label{Appendix: Synthetic Data}
We simulate two synthetic covariate-TPP datasets: covariate inhomogeneous Poisson process and covariate Hawkes process. Each of them has 1280 sequences, divided into training, validation and testing sets with an 8:1:1 ratio. Note that we run each model under different hidden state configurations: $M=M_V=M_K=[64, 128, 256, 512]$ with $\tilde{H}=H=4$, and report the best performance. For the covariate inhomogeneous Poisson process, the intensity is designed as:
\begin{equation*}
\lambda(t) = \mathbf{w}_{t}\mathbf{x}_n,
\end{equation*}
where $\mathbf{x}_n$ is the last covariate before $t$, and $\mathbf{w}_t$ is the weight to transform $\mathbf{x}_n$ to $\lambda(t)$. Additionally, the next event type is also related to the covariate, which can be specified as:
\begin{equation*}
    \begin{gathered}
    v_{n+1} = \underbrace{\mathbf{w}_c \mathbf{x}_n}_\text{covariate effect} + \underbrace{\frac{1}{n}\sum_{i=1}^n w_\tau                   (\tau_i)}_\text{history affect}, \\ 
    y_{n+1} =\left\{\begin{array}{l} 0 \text { if }  v_{n+1} \leq \zeta \\
                                    1 \text { if } v_{n+1} > \zeta
    \end{array}\right., 
    \end{gathered}
\end{equation*}
where $v_{n+1}$ is the logit generated by current covariate and historical information. $\tau_i=t_i-t_{i-1}$ is the interval between two adjacent events. The event type $y_{n+1}$ is set to type 1 when $v_{n+1}$ is larger than a predefined threshold $\zeta$, otherwise, it will be type 0. Note that the weight $\mathbf{w}_c$ is the ground-truth feature importance to be learned.
 
For the covariate Hawkes process, the covariate is also integrated into the intensity function, which is designed as: 
\begin{equation*}
\lambda(t) = \underbrace{\mathbf{w}_t \mathbf{x}_{n}}_\text{covariate effect} + \sum_{t_i<t} \alpha \exp(-\beta (t - t_i)), 
\end{equation*}
where, for simplicity, we assume the baseline intensity depends on the covariate and the triggering kernel is an exponential decay. The event type simulation in the covariate Hawkes process follows the same setting as the covariate inhomogeneous Poisson process. The detailed summary of our synthetic datasets is provided in~\cref{table: Synthetic Description}. 

\begin{table*}[t]
    \centering
    \caption{The statistics of two synthetic datasets are presented. We simulate covariate-based inhomogeneous Poisson processes and Hawkes processes. Each dataset contains 1280 sequences. Details such as the number of events, sequence length and the percentage of various event types are provided below.}
    \begin{sc}
    \scalebox{0.9}{
    \begin{tabular}{cccccccc}
        \toprule
        \multirow{2}{*}{Dataset} & \multirow{2}{*}{Split} & \multirow{2}{*}{\# of Events} & \multicolumn{3}{c}{Sequence Length} & \multicolumn{2}{c}{Event Type}\\
            &  &  & Max & Min & Mean(Std) & 0 & 1\\
        \midrule
        \multirow{3}{*}{Inhomogeneous} & training & 26505 & 43 & 11 & 25.88(5.03) & 52.59\%  & 47.41\% \\
            & validation & 3259 & 40 & 13 & 25.46(5.35) & 51.33\% & 48.67\% \\
            & test & 3275 & 37 & 16 & 25.59(4.51) & 51.63\% & 48.37\% \\
        \midrule
        \multirow{3}{*}{Hawkes} & training & 13692 & 75 & 1 & 13.37(11.74) & 68.08\% & 31.92\% \\
            & validation & 1699 & 56 & 2 & 13.16(10.69) & 70.87\% & 29.13\% \\
            & test & 1685 & 69 & 1 & 13.16(10.69) & 70.92\% & 29.08\% \\
        \bottomrule
    \end{tabular}}
    \end{sc}
    \label{table: Synthetic Description}
\end{table*}

\section{Real Data}
\label{Appendix: Real Data}
\begin{table*}[t]
\caption{The covariates used in two real datasets. (a): Eight meteorological features in the PM2.5 dataset as covariates. (b): Nine meteorological features in the London car accident dataset as covariates.} 
\begin{center}
\adjustbox{valign=b}{
\begin{minipage}{0.35\textwidth}
\subcaption[]{Covariates in the PM2.5 dataset.}
\centering
\scalebox{0.91}{
    \begin{tabular}{cc}
        \toprule
        Feature & Comment \\
        \midrule
        PM2.5 & PM2.5 Concentration $(\mu g/m^3)$ \\
        DEWP & Dew Point (Celsius Degree) \\
        TEMP & Temperature (Celsius Degree) \\
        HUMI & Humidity (\%) \\
        PRES & Pressure (hPa) \\
        IWS & Cumulated Wind Speed ($m/s$) \\
        PREC & Hourly Precipitation ($mm$) \\ 
        IPREC & Cumulated Precipitation ($mm$) \\
        \bottomrule
    \end{tabular}}
\label{table: PM2.5 covariates}
\end{minipage}}
\adjustbox{valign=b}{
\begin{minipage}{0.58\textwidth}
\subcaption[]{Covariates in the car accident dataset.}
\centering
\scalebox{0.83}{
\begin{tabular}{cc}
        \toprule
        Feature & Comment \\
        \midrule
        cloud cover & cloud cover measurement in oktas  \\
        sunshine & sunshine measurement in hours (hrs)  \\
        global radiation & irradiance measurement in Watt per square meter (W/m2) \\
        max temp & maximum temperature recorded in degrees Celsius (°C)  \\
        mean temp & mean temperature in degrees Celsius (°C)  \\
        min temp & minimum temperature recorded in degrees Celsius (°C) \\
        precipitation & precipitation measurement in millimeters (mm) \\ 
        pressure & pressure measurement in Pascals (Pa) \\
        snow depth & snow depth measurement in centimeters (cm)\\
        \bottomrule
    \end{tabular}}
\label{table: london-accident covariates}
\end{minipage}}
\end{center}
\label{table: dataset covariates}
\end{table*}
In this section, we present some statistics of the 2 public datasets: PM2.5 and London car accident datasets.
\subsection{PM2.5 dataset}
\label{Appendix: Real-Data PM2.5}
In this section, we present the PM2.5 dataset, which includes air pollutant recordings of PM2.5 concentration and meteorological data from five cities in China. The dataset spans from 2010 to 2015, and hourly meteorological records are considered as covariates that may impact the generalization of air pollutant alerts. Specifically, air pollutant alerts are triggered when the PM2.5 concentration exceeds a certain threshold, which is the event of interest in our study. Following the EPA Victoria AirWatch standard, four types of events are recognized: Fair Alert (PM2.5 rising to $25 \mu g/m^3$ from a lower value), Poor Alert (PM2.5 rising to $50 \mu g/m^3$), Very Poor Alert (PM2.5 rising to $100 \mu g/m^3$), and Extremely Poor Alert (PM2.5 rising to $300 \mu g/m^3$). We train different models for each city and split the dataset into training (2010-2013), validation (2014), and testing (2015) sets. The statistical information of each split for each city is provided in \cref{table: PM2.5 Data Description}. 
\begin{table*}[th]
    \centering
    \caption{The statistics of the PM2.5 Alert dataset are presented, featuring six years of PM2.5 data from five cities. The training data includes information from 2010, 2011, 2012, and 2013, while the validation set comprises 2014 data and the testing set contains 2015 data.}
    \scalebox{0.9}{
    \begin{sc}
    \begin{tabular}{cccccccc}
        \toprule
        \multirow{2}{*}{Dataset} & \multirow{2}{*}{Split} & \multirow{2}{*}{\# of Readings} & \multicolumn{4}{c}{Event \%} & \multirow{2}{*}{Avg Interval}\\
         &  &  & Fair & Poor & VP & EP & \\
        \midrule
        \multirow{3}{*}{Beijing} 
            & training & 41756 & 27.28\% & 33.07\% & 32.83\% & 6.83\% & 11.61 \\
            & validation& 8661 & 27.27\% & 32.15\% & 34.65\% & 5.93\% & 11.52 \\
            & testing & 8630 & 33.57\% & 33.01\% & 26.74\% & 6.69\% & 12.24 \\
        \midrule
        \multirow{3}{*}{Shenyang} 
            & training & 13774 & 31.43\% & 39.66\% & 26.99\% & 1.92\% & 8.12 \\
            & validation& 8393 & 27.65\% & 40.05\% & 30.33\% & 1.96\% & 7.81 \\
            & testing & 7906 & 33.54\% & 41.06\% & 22.33\% & 3.07\% & 9.28 \\
        \midrule
        \multirow{3}{*}{Shanghai} 
            & training & 25707 & 44.86\% & 37.76\% & 16.80\% & 0.58\% & 10.88 \\
            & validation& 8615 & 48.39\% & 37.00\% & 14.36\% & 0.25\% & 10.82 \\
            & testing & 8332 & 49.93\% & 38.54\% & 11.39\% & 0.13\% & 11.59 \\
        \midrule
        \multirow{3}{*}{Guangzhou} 
            & training & 23836 & 39.46\% & 41.47\% & 18.87\% & 0.20\% & 11.16 \\
            & validation& 8077 & 37.30\% & 46.34\% & 16.10\% & 0.26\% & 11.50 \\
            & testing & 8516 & 54.42\% & 36.58\% & 9.00\% & 0.00\% & 13.13 \\
        \midrule
        \multirow{3}{*}{Chengdu} 
            & training & 20251 & 19.14\% & 46.01\% & 32.55\% & 2.31\% & 12.06 \\
            & validation& 8475 & 20.30\% & 49.47\% & 29.17\% & 1.05\% & 10.20 \\
            & testing & 8649 & 26.87\% & 47.18\% & 25.43\% & 0.52\% & 11.56 \\
        \bottomrule
    \end{tabular}
    \end{sc}}
    \label{table: PM2.5 Data Description}
\end{table*}
\begin{table*}[th]
    \centering
    \caption{The statistics of the London car accident dataset are presented, featuring 10 years of car accidents data from London during the winter season. The training data includes information from 2005~2012, while the validation set comprises 2013~2014 and the testing set contains 2015 data.}
    \scalebox{0.9}{
    \begin{sc}
    \begin{tabular}{ccccccc}
        \toprule
        \multirow{2}{*}{Dataset} & \multirow{2}{*}{Split} & \multirow{2}{*}{\# of Readings} & \multicolumn{3}{c}{Event \%}& \multirow{2}{*}{Avg Interval}\\
         &  &  & Fatal & Medium & Slight & \\
        \midrule
        \multirow{3}{*}{London-Car-Accident} 
            & training & 26331 & 4.77\% & 32.34\% & 62.88\% & 53.2  \\
            & validation & 9212 & 3.28\% & 31.12\% & 65.60\%& 47.1\\
            & testing & 4281 & 2.62\% & 34.48\% & 62.90\%& 48.9\\
        \bottomrule
    \end{tabular}
    \end{sc}
}
    \label{table: London car-accident Data Description}
\end{table*}
\subsubsection{Extra Results: Guangzhou, Shanghai Chengdu}
This is the additional experiment about the PM2.5 Alert dataset, presenting the results of other 3 cities: Shanghai, Chengdu and Guangzhou. Similar conclusions can be drawn that, the enhanced models (RMTPP+, THP+) outperform the original ones; and our proposed TransFeat achieves competitive performance compared with others.

\begin{table*}[tb]
\centering
\caption{Experimental results on the other 3 cities: Guangzhou, Shanghai, Chengdu. It can be drawn the similar conclusions that, the enhanced model, RMTPP+ and THP+, outperform the original models, proving that incorporating covariates can enhance the model's predictive performance. And our proposed TransFeat-TPP achieves the best performance in most cases.} 
\label{table: Extra Experiment Result}
\begin{sc}
\scalebox{0.79}{
\begin{tabular}{ccccc|cccc|cccc} 
\toprule
\multirow{2}{*}{Model} & \multicolumn{4}{c}{PM2.5-Guangzhou} & \multicolumn{4}{c}{PM2.5-Shanghai} & \multicolumn{4}{c}{Car-accident-Chengdu}\\
\cmidrule{2-13}
    Model & LOG-LL & RMSE & ACC & F1 & LOG-LL & RMSE & ACC & F1 & LOG-LL & RMSE & ACC & F1\\ 
\midrule

    \multirow{2}{*}{RMTPP} & -0.75 & 0. 22 & 61.55\% & 0.58 & -1.87 & 0.34 & 57.75\% & 0.55 & -4.58 & 1.31 & 55.50\% & 0.49\\
    & (0.010)&(0.032) & (0.009)& (0.011)& (0.029)&(0.045) & (0.012)& (0.005) & (0.028)&(0.042) & (0.018)& (0.029) \\
    
    \multirow{2}{*}{RMTPP+} & -0.74 & 0.21 & \textbf{65.78}\% & 0.61 & -1.90 & 0.32 & 65.24\% & 0.58 & -4.66 & 1.26 & 63.40\% & 0.58\\
    & (0.015) & (0.055) & (0.018) & (0.022) & (0.020) & (0.056) & (0.011) & (0.009) & (0.095) & (0.080) & (0.054)& (0.031) \\
    
    \multirow{2}{*}{RNNPP} & --- & 0.28 & 53.10\% & 0.5918 & --- & 0.38 & 0.61 & \textbf{65.89\%} & --- & 1.13 & 0.68 & 0.68\\
       & --- & (0.010) &(0.015) & (0.015)& --- & (0.022) &(0.020) &(0.021)& --- &(0.012) &(0.016) & (0.019)  \\
    
    \multirow{2}{*}{THP} & -0.49 & 0.22 & 58.90\% & 0.55 & -2.68 & 0.27 & 0.54 & 57.76\% & -4.12 & 1.20 & 60.03\% & 0.56\\
        & (0.051)&(0.098) & (0.017)& (0.078)& (0.097)& (0.010) & (0.043)& (0.014) & (0.069)&(0.050) & (0.044)& (0.050) \\
    
    \multirow{2}{*}{THP+} & -0.38 & \textbf{0.19} & 61.32\% & 0.58 & -2.51 & 0.28 & 59.90\% & 0.54  & -4.32 & 1.07 & 65.03\% & 0.63\\
        & (0.069)& (0.085) & (0.023) & (0.085) & (0.120) & (0.031) & (0.031) & (0.071) & (0.042) & (0.013) & (0.051)& (0.039) \\
    
     \multirow{2}{*}{\textbf{TransFeat}} & \textbf{-0.29} & 0.23 & 64.89\% & \textbf{0.63} & \textbf{-1.20}  & \textbf{0.17} & \textbf{68.39\%} & 0.64 & \textbf{-3.10} & \textbf{0.97} & \textbf{71.33\%} & \textbf{0.72}\\
         & (0.055)&(0.049) & (0.081)& (0.082)& (0.088)& (0.031) & (0.059)& (0.010) & (0.054)&(0.043) & (0.037)& (0.046) \\
\bottomrule
\end{tabular}}
\end{sc}
\end{table*}
\subsection{London car accident dataset}
\label{Appendix: Real-Data London-Car-Accident}
The dataset utilized in this study originates from UK police forces and encompasses records of vehicle collisions that took place in the UK between 2005 and 2015. The primary focus of the investigation is on car accidents occurring during the winter season in London. Each sequence represents accidents that occur on a daily basis, and their corresponding severity levels are indicated as markers. More details are shown in~\cref{table: London car-accident Data Description}. Besides, we incorporate the London metgeroical data to investigate their correlation to car accident severity, which includes 9 features: cloud cover, sunshine, global radiation, maximum temperature, minimum temperature, precipitation, pressure and snow depth.
\section{Log-normal Mixture Model}
\label{app_log_normal_mixture}
In this section, we provide details about the log-normal mixture distribution of inter-event time in the decoder module. 
Specifically, we define the log-normal mixture distribution as follows: 
\begin{equation*}
\begin{gathered}
z_1 \sim \text{GaussianMixture}(\mathbf{w}, \mathbf{s}, \bm{\mu}), 
\\
z_2 = az_1 + b,
\\
\tau = \exp(z_2), 
\\
\end{gathered}
\end{equation*}
where the affine transformation $z_2=az_1+b$ can facilitate better initialization according to \cite{shchur2019intensity}, and $\tau$ is the inter-event time.
We need to compute two outputs: (1) the log-likelihood, which is part of the objective function; (2) the predicted next event time, which is the expectation of our log-normal mixture distribution.

\subsection{Log-likelihood}
Given the historical information up to $t_i$, we can obtain the latent representation $\mathbf{h}_i$, and then we can compute the parameters: $\{\mathbf{w}_i, \mathbf{s}_i, \bm{\mu}_i \}$, so we can compute the probability density function of $\tau_{i+1} = t_{i+1} - t_{i}$: 
\begin{equation*}
    \begin{gathered}
        \frac{\log\tau_{i+1}-b}{a} \sim  \text{GaussianMixture}(\mathbf{w}_i, \mathbf{s}_i, \bm{\mu}_i),\\
        p(\tau=\tau_{i+1}) = f_i( \frac{\log\tau_{i+1}-b}{a}), 
    \end{gathered}
\end{equation*}
where $f_i$ is the probability density function of Gaussian mixture distribution with parameters: $\{ \mathbf{w}_i, \mathbf{s}_i, \bm{\mu}_i \}$. Therefore, the overall log-likelihood can be computed as: 
\begin{equation*}
    \begin{gathered}
        \mathcal{L} = \sum_{i=0}^{L-1} \log f_{i}( \frac{\log\tau_{i+1}-b}{a}). 
    \end{gathered}
\end{equation*}

\subsection{Time prediction}
To predict the next event's timestamp, we need to compute the expectation of the log-normal mixture distribution. \cite{shchur2019intensity} provides an analytical expression to compute this expectation value: 
\begin{equation*}
    \begin{aligned}
        \bar{\tau} = \sum_k w_k \exp (a\mu_k + b + \frac{(as_k)^2}{2}). 
    \end{aligned}
\end{equation*}

\end{document}